\documentclass{article}

\pdfoutput=1

\usepackage[nonatbib, final]{svrhm_2022}

\usepackage[utf8]{inputenc} 
\usepackage[T1]{fontenc}    
\usepackage{hyperref}       
\usepackage{url}            
\usepackage{booktabs}       
\usepackage{amsfonts}       
\usepackage{nicefrac}       
\usepackage{amsmath}        
\usepackage{microtype}      
\usepackage{graphicx}       %
\usepackage{svg}
\usepackage{subcaption}

\usepackage{tikz}           
\usepackage{wrapfig}        
\usepackage[export]{adjustbox} 
\usepackage{flafter}        
\usepackage{floatrow}       
\usepackage{appendix}       

\usepackage{xcolor}         

\usepackage[backend=biber,sorting=none]{biblatex}
\usepackage{chngcntr}

\bibliography{references}
\title{Connecting metrics for shape-texture knowledge in computer vision}

\author{%
  Tiago Oliveira$^{1}$, Tiago Marques$^{2,3,4,5}$,  Arlindo L. Oliveira$^{1,6}$\\
  \\
  $^1$Instituto Superior Tecnico / INESC-ID, Lisbon, Portugal  \\
  $^2$Department of Brain and Cognitive Sciences, MIT, Cambridge, MA \\
  $^3$McGovern Institute for Brain Research, MIT, Cambridge, MA \\
  $^4$Center for Brains, Minds and Machines, MIT, Cambridge, MA \\
  $^5$Champalimaud Clinical Centre, Champalimaud Foundation, Lisbon, Portugal \\
  $^6$Computer Science and Artificial Intelligence Laboratory, MIT, Cambridge, MA
  \\
}

\begin{document}
\maketitle

\begin{abstract}
Modern artificial neural networks, including convolutional neural networks and vision transformers, have mastered several computer vision tasks, including object recognition.  However, there are many significant differences between the behavior and robustness of these systems and of the human visual system. Deep neural networks remain brittle and susceptible to many changes in the image that do not cause humans to misclassify images. Part of this different behavior may be explained by the type of features humans and deep neural networks use in vision tasks. Humans tend to classify objects according to their shape while deep neural networks seem to rely mostly on texture. Exploring this question is relevant, since it may lead to better performing neural network architectures and to a better understanding of the workings of the vision system of primates. In this work, we advance the state of the art in our understanding of this phenomenon, by extending previous analyses to a much larger set of deep neural network architectures. We found that the performance of models in image classification tasks is highly correlated with their shape bias measured at the output and penultimate layer. Furthermore, our results showed that the number of neurons that represent shape and texture are strongly anti-correlated, thus providing evidence that there is competition between these two types of features. Finally, we observed that while in general there is a correlation between performance and shape bias, there are significant variations between architecture families.
  
\end{abstract}

\section{Introduction}
\label{introduction}

%

Modern artificial neural networks (ANNs) such as some convolutional neural networks (CNNs) \cite{tan2019efficientnet,xie2020self,liu2022convnet} and vision transformers \cite{dosovitskiy2020image,liu2021swin} have achieved human-level performance in challenging computer vision tasks like object recognition. Additionally, some of these models often learn representations that resemble those found in neuronal populations along the primate visual ventral stream areas \cite{zhuang2021unsupervised} suggesting a parallelism in how visual information is processed in both ANNs and biological neural networks in the brain. However, significant differences between these systems remain, namely in the ANNs' vulnerability to adversarial examples \cite{szegedy2014intriguing, kurakin2018adversarial}, or their lack of generalization capacity to out-of-distribution (OOD) datasets \cite{geirhos2018generalisation, hendrycks_benchmarking_2019}. Another such difference is in the type of features humans and ANNs use predominantly for making inferences. Unlike humans, who tend to classify objects according to their shape, ANNs have been reported to rely mostly on texture cues \cite{geirhos2018imagenet, hermann2020origins}. Due to this, ANNs are known for exhibiting a 'texture bias' or lacking a more human-aligned 'shape bias'. These biases are typically studied using images with conflicting texture and shape cues while recording the ANNs' behavioral outputs \cite{geirhos2018imagenet, hermann2020origins} or their internal activity \cite{islam2021shape}. 
 
 
 There are multiple reasons why exploring this question is relevant. From a neuroscientific perspective, there is an increasing interest in using ANNs to model brain processing, particularly visual behaviors such as object recognition. Therefore, the apparent lack of shape bias of ANNs is a crucial limitation that should be addressed. From a machine learning standpoint, understanding these divergences might provide insights on how to build networks that are more aligned with human vision, and thus more interpretable, as well as more accurate and robust. 
 Multiple studies have related robustness, OOD generalization and shape bias, but a coherent theory connecting them has not emerged, yet. Reasoning using shape cues was initially thought to explain the success of human vision to generalize to many domains \cite{geirhos2018imagenet}, though more recent studies have contradicted this hypothesis \cite{hermann2020origins, mummadi2021does}. On the other hand, there is plenty of evidence relating adversarial robustness with increased shape bias \cite{geirhos2021partial, chen2020shape, zhang2019interpreting}). Interestingly, the relationship between adversarial training and OOD generalization accuracy is less clear with some studies suggesting some degree of correlation \cite{xie2020adversarial} and others little connection \cite{dapello_simulating_2020}. Another interesting result is that while maximizing accuracy during standard training leads to increases in model shape-bias, the opposite is not necessarily true \cite{hermann2020origins}.
 
In this work, we build on and complement existing literature on this subject to fill in existing gaps. Hermann et al. showed that the shape bias measured at the ANNs outputs (behavior) was correlated with their ImageNet accuracy \cite{hermann2020origins}. Similarly, Islam et al. used a metric to estimate the shape dimensionality bias in latent representations of ANNs and found that it was also related with their accuracy \cite{islam2021shape}. Both these studies used a very small number of models with little diversity in their architectures to make these inferences. The contributions of our work are the following:

\begin{enumerate}
    \item We extend and combine the results of Hermann et al. and Islam et al. to a much larger pool of models (more than 600) with considerably different network architectures (both CNNs and vision transformers). We found that as models achieve higher ImageNet accuracy, they also increase their shape bias measured at the output and in their penultimate layer.
    \item We observed that shape  and texture dimensionalities are strongly anti-correlated, thus quantifying the trade-off between the model's knowledge of these two concepts.
    \item We studied these relationships in the context of different model families, observing a large variance in the results despite the overall correlations. 
\end{enumerate}



\section{Methods}
\label{methods}
\subsection{Shape Bias Evaluation}
\subsubsection{Shape Behavioral Bias}
For a dataset that has a shape-texture conflict in each image (for example, an image with a cat shape and dog texture), the shape behavioral bias, here termed \textit{shape-bias}, measures the inclination of the model to classify images according to shape, as opposed to texture \cite{geirhos2018imagenet}. This is given by:
\begin{equation}
\text{shape-bias} = \frac{\# \text{correct shapes}}{\# \text{correct shapes} + \# \text{correct textures}}
\end{equation}

\subsection{Shape Dimensionality Bias}
Following the definition of Esser et al., we here use the term dimensionality to quantify the number of neurons associated with the \textit{knowledge} about a given concept in a neural network layer \cite{esser2020disentangling}. 
In this study, we consider three concepts, or \textit{factors}: \textbf{shape}, \textbf{texture} and \textbf{residual}. The latter captures all the remaining knowledge in a given layer that is neither shape or texture. 

In order to estimate the dimensionality of a certain factor, multiple pairs of images,  $x^a$ and $x^b$, are presented to an ANN, and a given layer's activations, respectively $z^a$ and $z^b$, are collected. For each image pair, the factor of interest is shared while the other differs (for example, two images of the same object with one having its texture swapped to that of another object). The idea behind this approach is to estimate how much these activations vary if the factor of interest is kept. To do this, we measure how much do $z^a$ and $z^b$ correlate (neuron-wise),
\begin{equation}
\rho_\text{factor} = \frac{1}{N} \sum_i^N \frac{\text{Cov}(z_i^a, z_i^b)}{\sqrt{\text{Var}(z_i^a)\text{Var}(z_i^b)}},
\end{equation}
where $i$ is the neuron number and $\rho_\text{residual}$ is $1$ by definition. The stronger the correlation, the more relevant that factor is for that model layer's representation. To obtain the final estimate of dimensionality of the factor, we then apply the softmax function to all factor correlations and multiply them by the total number of neurons in the layer. Here, we employ this approach to study the dimensionality of these factors in the penultimate model layer. Unlike Islam et al. \cite{islam2021shape}, we propose to study the shape-texture bias, not with the absolute dimensionality, but with the fraction between shape and texture neurons which we call \textit{shape-dim-ratio},
\begin{equation}
\text{shape-dim-ratio} = \frac{\text{shape-dim}}{\text{shape-dim} + \text{texture-dim}}
\end{equation}

\subsection{Visual Stimuli}
We use the cue-conflict image dataset proposed by Geirhos et al. to calculate the shape behavioral bias in ANNs \cite{geirhos2018imagenet}. This dataset contains images generated with neural style transfer \cite{gatys2016image}. Each generated image combines the shape of one original image with the texture of another. In total, there are 16 object classes, with 10 examples per each class providing the shape, and 3 examples of each class providing the texture (1200 generated images). For each image, the shape and texture class are never the same - this is the cue-conflict.

For the shape dimensionality bias estimate, we use Stylized Pascal VOC 2012 \cite{islam2021shape}, which was created by combining a subset of the PASCAL VOC 2012 dataset \cite{everingham2010pascal} and the Describable Textures Dataset \cite{cimpoi2014describing}. Each object image (from 20 classes) is stylized \cite{gatys2016image} into 5 new ones, using 5 different texture images (21845 generated images). This allows to sample image pairs that are either similar in shape or similar in textures.

\subsection{Models}
We consider a very large pool of ANN models (624 models trained on ImageNet) to measure both the shape behavioral and dimensionality biases in order to better analyze and compare these metrics.  These models vary greatly in their architecture in terms of their overall structure (CNNs and vision transformers), architectural families, number of layers, data augmentation, ImageNet accuracy, and so on. 

For one analysis, we look in more detail in how the results change for different model families. We chose all the families that had at least nine different models, forming a total of 17 families:
Cait \cite{touvron2021going} (n=10), convnext \cite{liu2022convnet} (n=15), crossvit \cite{chen2021crossvit} (n=11), deit \cite{touvron2021training} (n=22), efficientnet \cite{tan2019efficientnet} (n=38), efficientnetv2 \cite{tan2021efficientnetv2} (n=11), hrnet \cite{wang2020deep} (n=9), mobilenet \cite{howard2017mobilenets, sandler2018mobilenetv2, howard2019searching} (n=17), mobilevit \cite{mehta2021mobilevit} (n=16), regnet \cite{xu2022regnet} (n=35), resmlp \cite{touvron2021resmlp} (n=9), resnet \cite{he2016deep} (n=89), swin \cite{liu2021swin} (n=9), swinv2 \cite{liu2022swin} (n=13), vit \cite{dosovitskiy2020image}, volo \cite{yuan2021volo} (n=10), and  xcit \cite{ali2021xcit} (n=42).


To measure the object classification performance of a given model, we use the reported top-1 accuracy on ImageNet (ILSVRC-2012 validation set). 

\section{Results}
\label{results}

\subsection{Shape Behavioral Bias, Shape Dimensionality Bias, and Accuracy}
\begin{figure}[h!]
    \begin{subfigure}{0.32\textwidth}
        \centering
        \includegraphics[width=\linewidth]{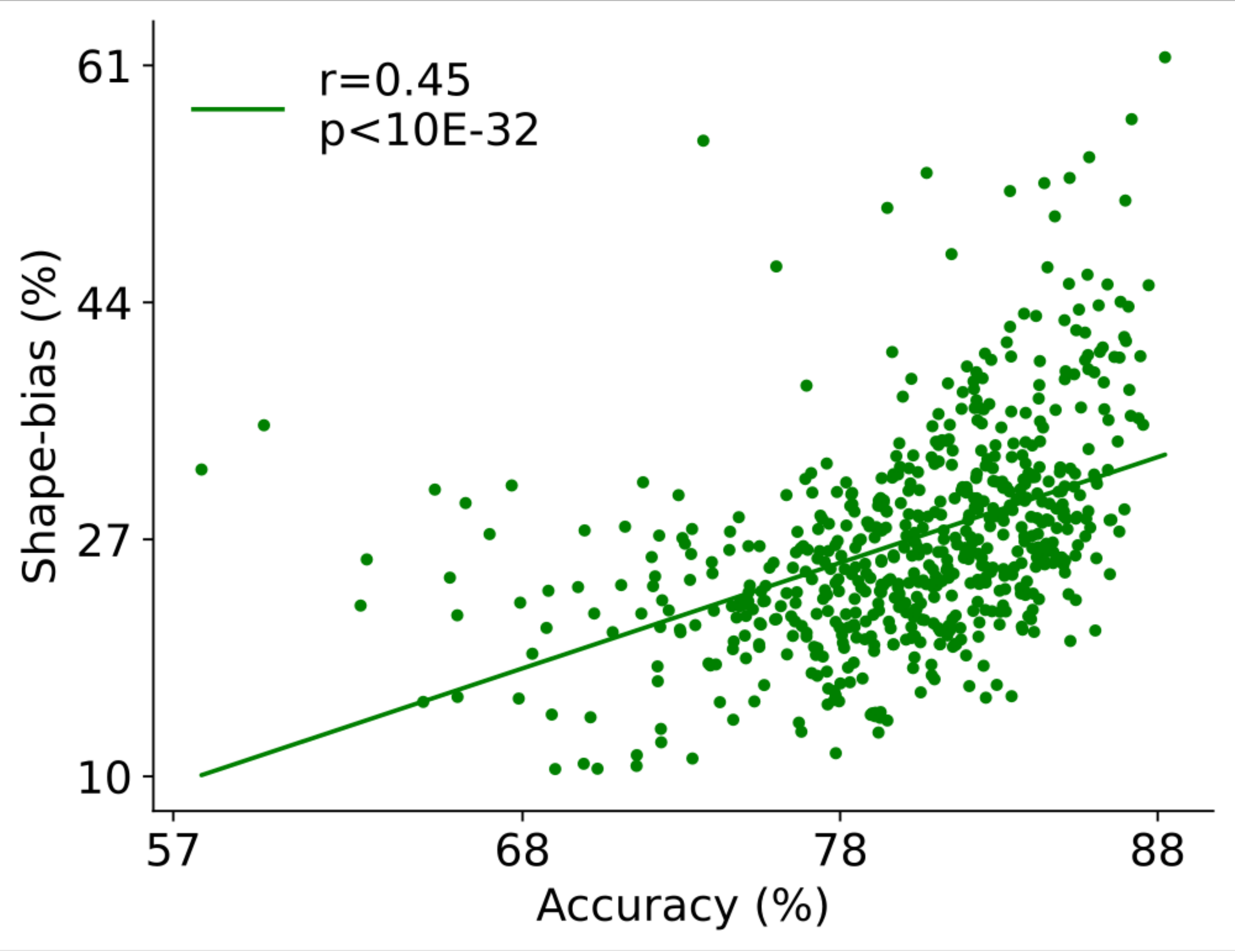}
    \end{subfigure}
    \begin{subfigure}{0.32\textwidth}
        \centering
        \includegraphics[width=\linewidth]{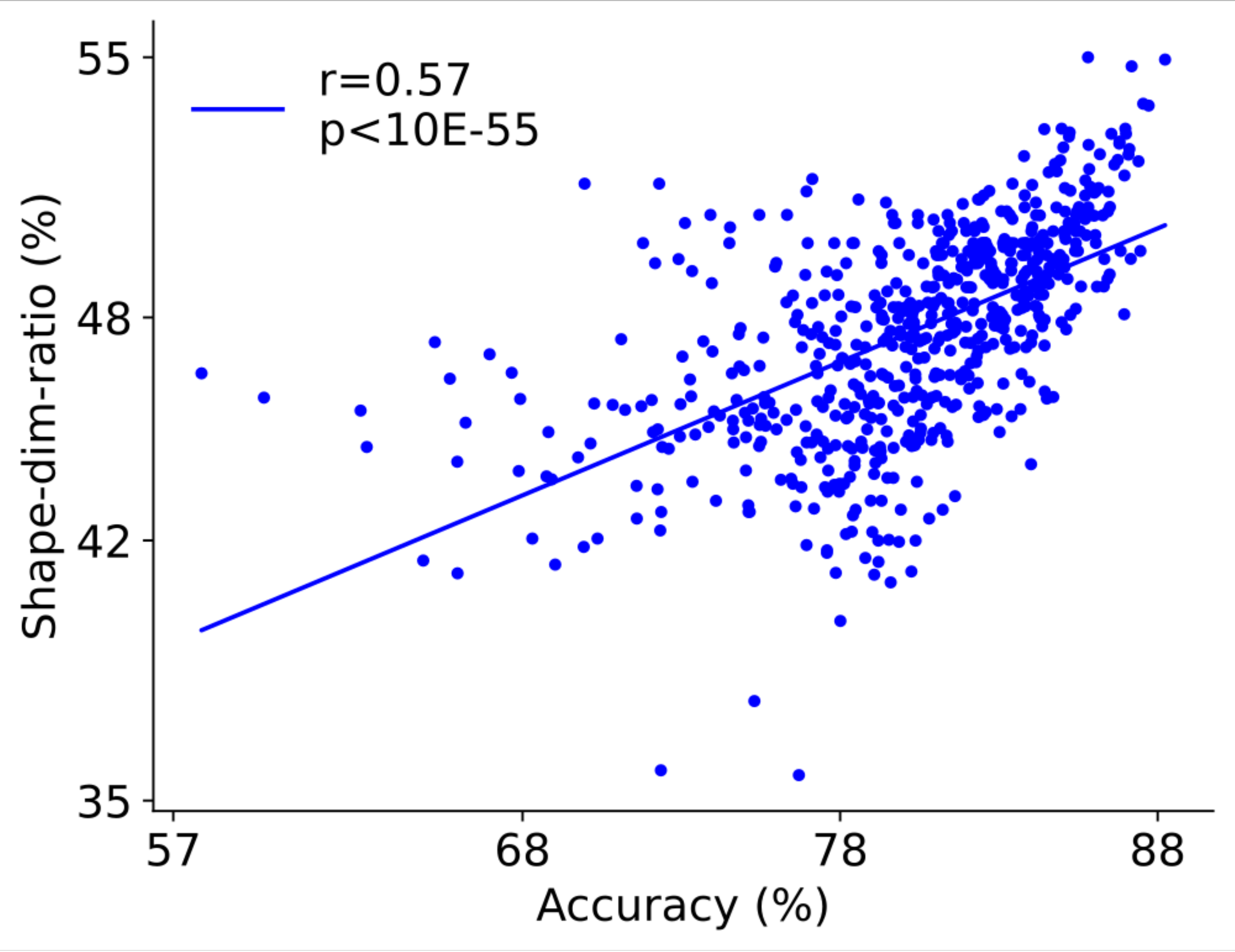}
    \end{subfigure}
    \begin{subfigure}{0.32\textwidth}
        \centering
        \includegraphics[width=\linewidth]{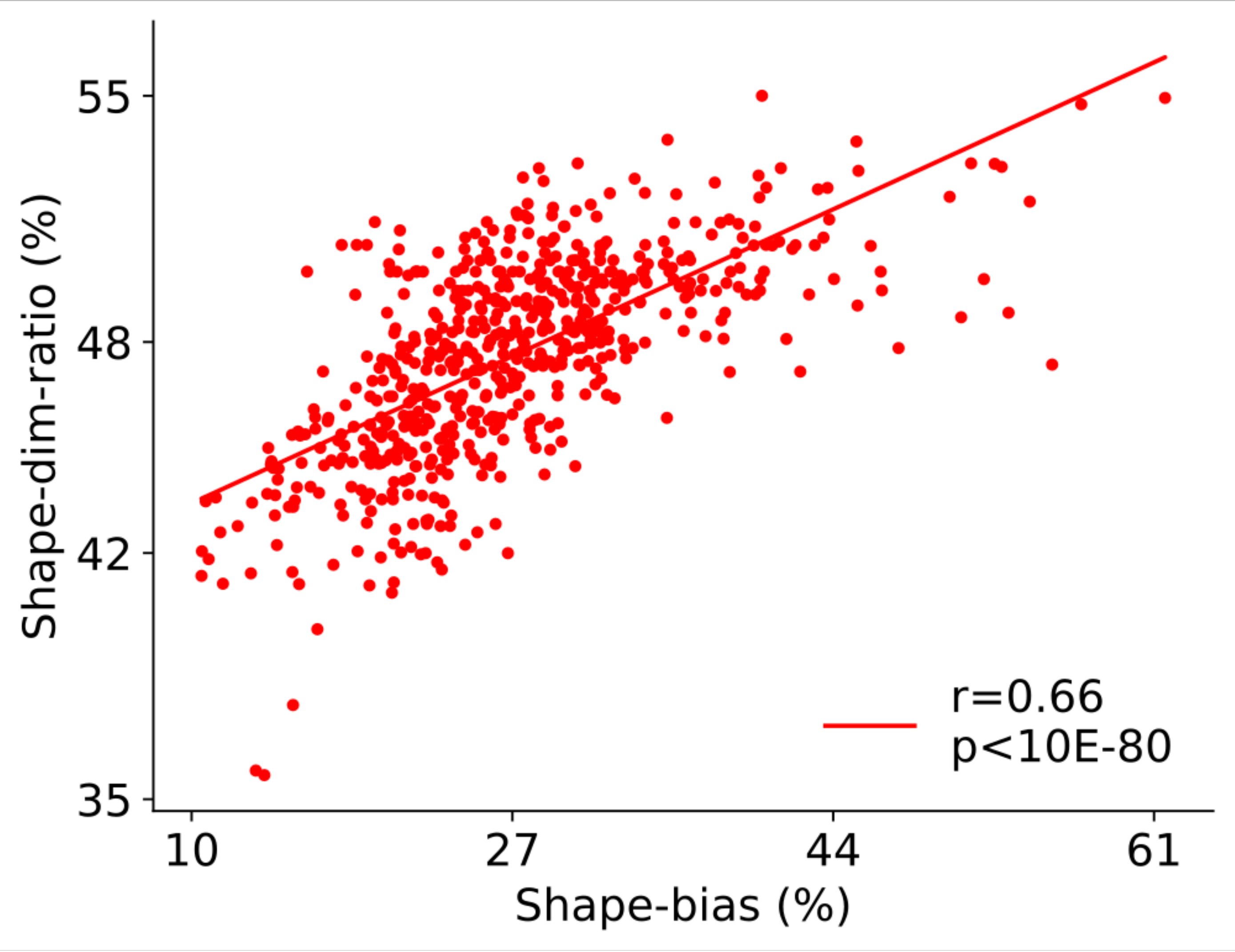}
    \end{subfigure}
    \caption{\textbf{Comparison of shape dimensionality metrics and model accuracy.} Shape behavioral bias (shape-bias), shape dimensionality bias (shape-dim-ratio), and object classification performance (ImageNet top-1 accuracy) are positively correlated. Model pool consists of 624 ImageNet trained models. r is the Pearson correlation coefficient.}
    \label{metrics_correlations}
\end{figure}

Similarly to the results in Herman et al. \cite{hermann2020origins}, we observed a positive correlation between the shape behavioral bias and object recognition performance in our model pool (Figure \ref{metrics_correlations}, left; r=0.45,p<10E-32). The models tested vary considerably more than those in the previous study in both these two metrics (accuracy $\sim$60\%-90\%, shape-bias $\sim$10\%-60\%), thus extending the range in which this relationship is observed.
While the correlation is far from perfect, our results show that the vast majority of models with relatively high shape behavioral bias (larger than 40\%) also have relatively high ImageNet accuracy (larger than 80\%).
Interestingly, we observe that some models solely trained for ImageNet performance without any optimization for shape bias can achieve very high shape bias. For example, one model (Noisy Student EfficientNet L2 \cite{xie2020self}) has a shape-bias larger than 60\%, something that previously had only been reported in models exploring more unconventional data augmentations, usually at the cost of ImageNet performance \cite{hermann2020origins}. This was not the case here, as the model was also the one with the highest object classification accuracy. 

Our model pool also varies considerably in the shape dimensionality bias (shape-dim-ratio $\sim$35\%-55\%). This variance is also correlated with ImageNet accuracy, and, similarly to the shape-dim, the models with the highest shape-dim-ratio were also the best performing models in  ImageNet (Figure \ref{metrics_correlations}, center; r=0.57, p<10E-55). 

Finally, when we directly compare the two shape bias metrics, we observe that the shape bias present at the behavioral output correlates strongly with the shape bias in the internal representations (Figure \ref{metrics_correlations}, right; r=0.66, p<10E-80). While this result was expected, it is important to emphasize that these metrics were measured using different image datasets. Interestingly, for models with very high shape-bias, further improving this metric leads to diminishing returns in shape-dim-ratio.

\subsection{Trade-Off between Shape and Texture Dimensionality}
\begin{figure}[H] 
    \begin{subfigure}{0.35\textwidth}
        \includegraphics[width=\linewidth]{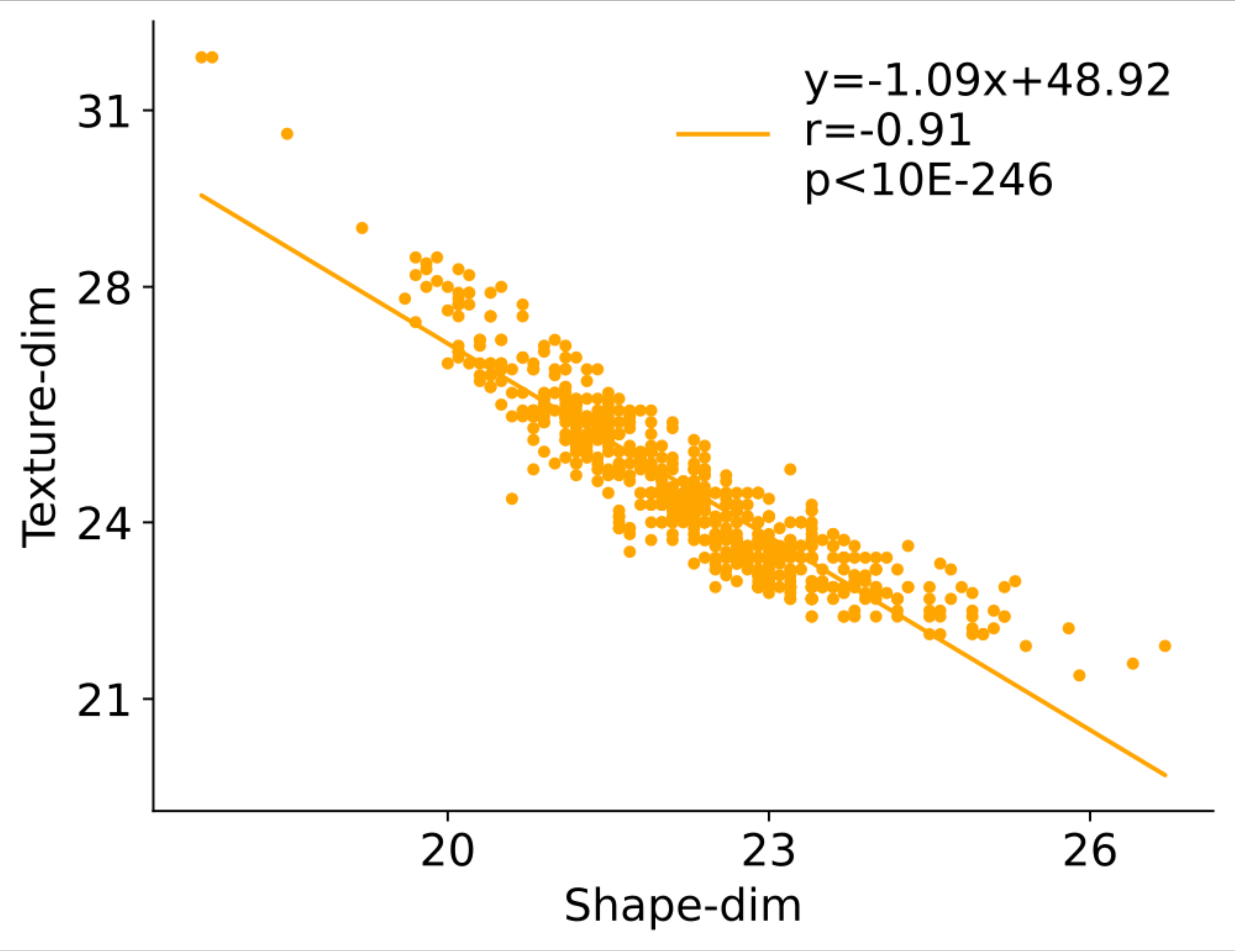}
    \end{subfigure}
    \begin{subfigure}{0.35\textwidth}
        \includegraphics[width=\linewidth]{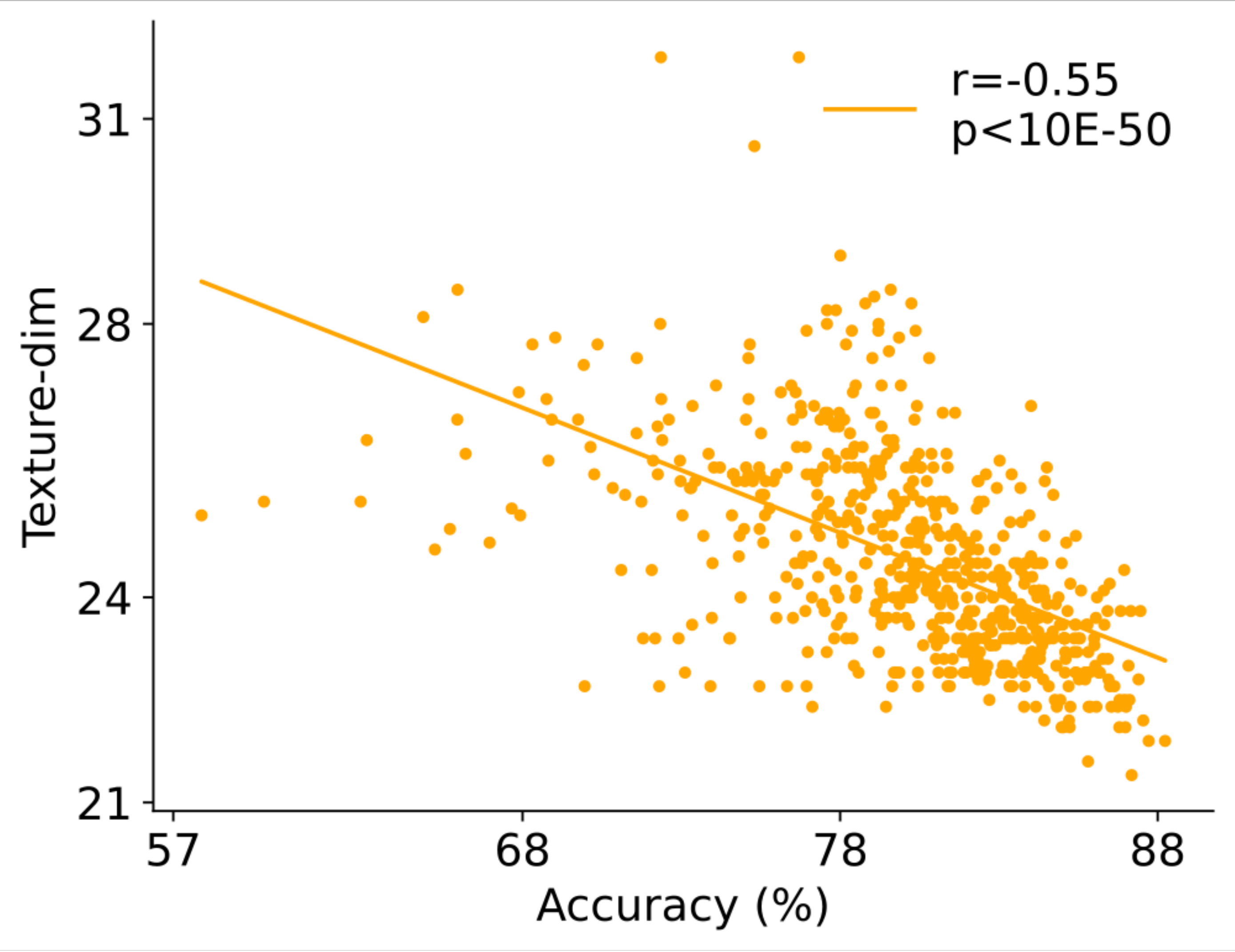}
    \end{subfigure}
    \caption{\textbf{Trade-off between shape and texture dimensionality in models' penultimate layer.} Left, texture and shape dimensionality are negatively correlated with a slope $\sim$-1. Right, as models perform better in object classification they have lower texture dimensionality.}
    \label{texture-dim_shape-dim}
\end{figure}

Previously, Islam et al., using a very small set of models, noticed that there was a trend for the shape dimensionality to decrease with increasing texture dimensionality. With the total dimensionality being capped at 100\%, there is a trade-off between the texture, shape and residual dimensionalities. Here, we quantify and measure this trade-off in our large model pool. Over the whole ranges of absolute shape dimensionality ($\sim$18\%-27\%) and texture dimensionality ($\sim$21\%-32\%), these two metrics are negatively correlated with a slope of approximately -1 (Figure \ref{texture-dim_shape-dim}, left; r=-0.91, p<10E-246). In other words, as models acquire more shape neurons, they lose texture neurons in the same proportion on average. Because of this, the number of residual neurons is similar across the model pool (residual dim $\sim$50\%-55\%, data not shown). Finally, another consequence of this trade-off is that with increasing performance in object recognition, models have smaller texture dimensionality (Figure \ref{texture-dim_shape-dim}, right; r=-0.55, p<10E-50).


\subsection{Correlation between Shape Bias Metrics and Accuracy across Architecture Families}

\begin{figure} [!h]
    \begin{subfigure}{0.38\textwidth}
        \includegraphics[width=\linewidth]{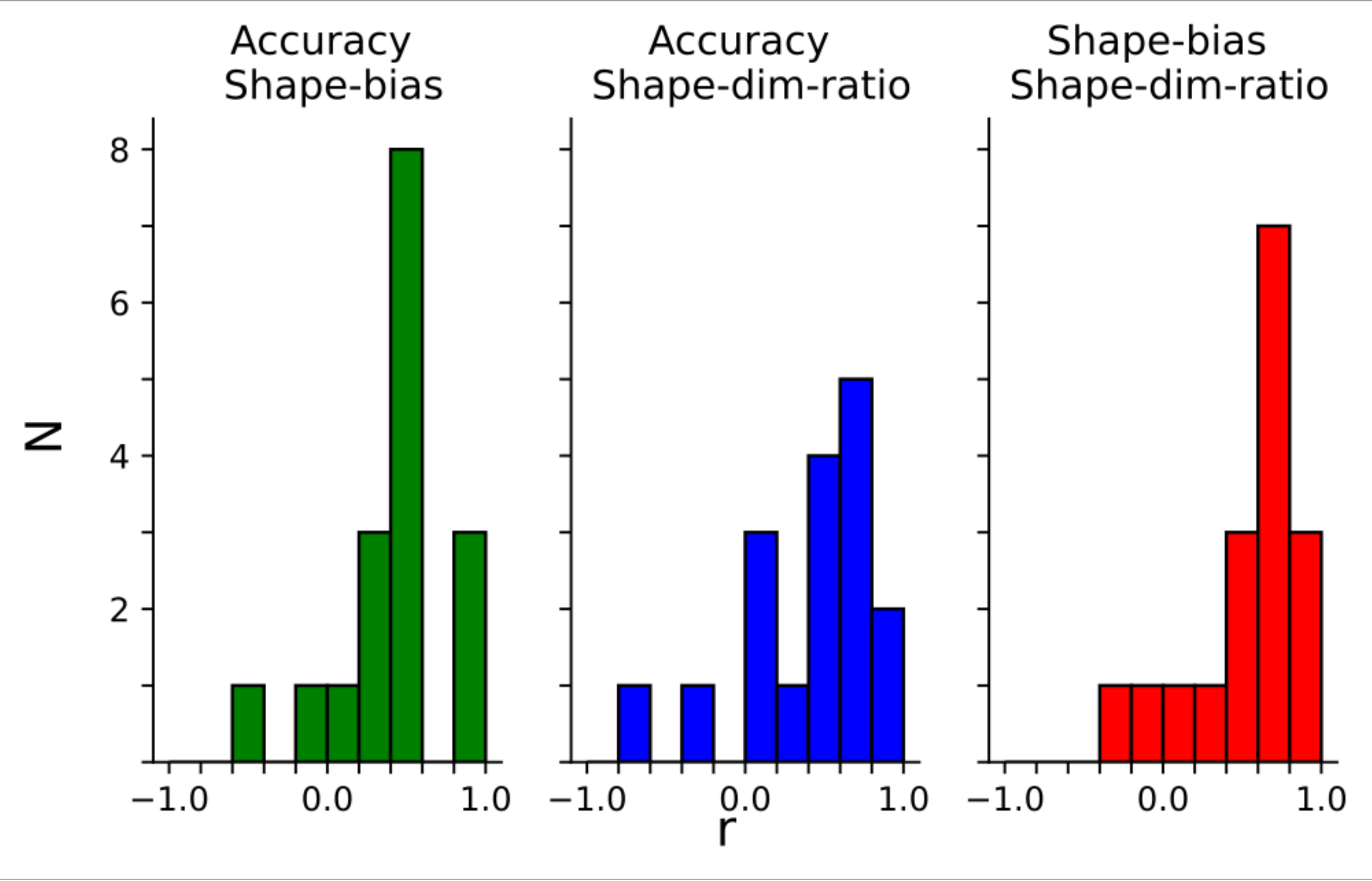}
    \end{subfigure}
    \begin{subfigure}{0.30\textwidth}
        \centering
        \includegraphics[width=\linewidth]{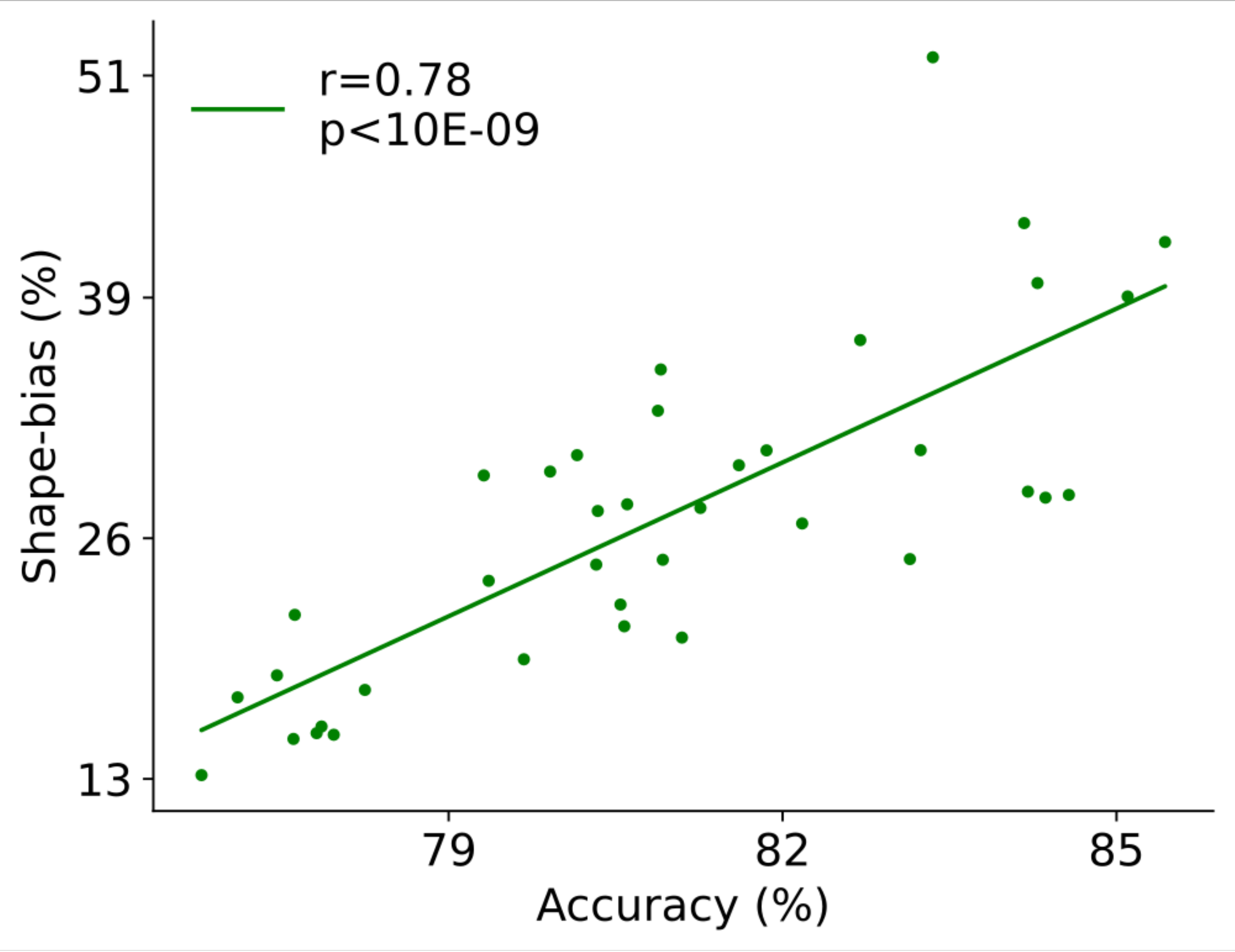}
    \end{subfigure}
    \begin{subfigure}{0.30\textwidth}
        \centering
        \includegraphics[width=\linewidth]{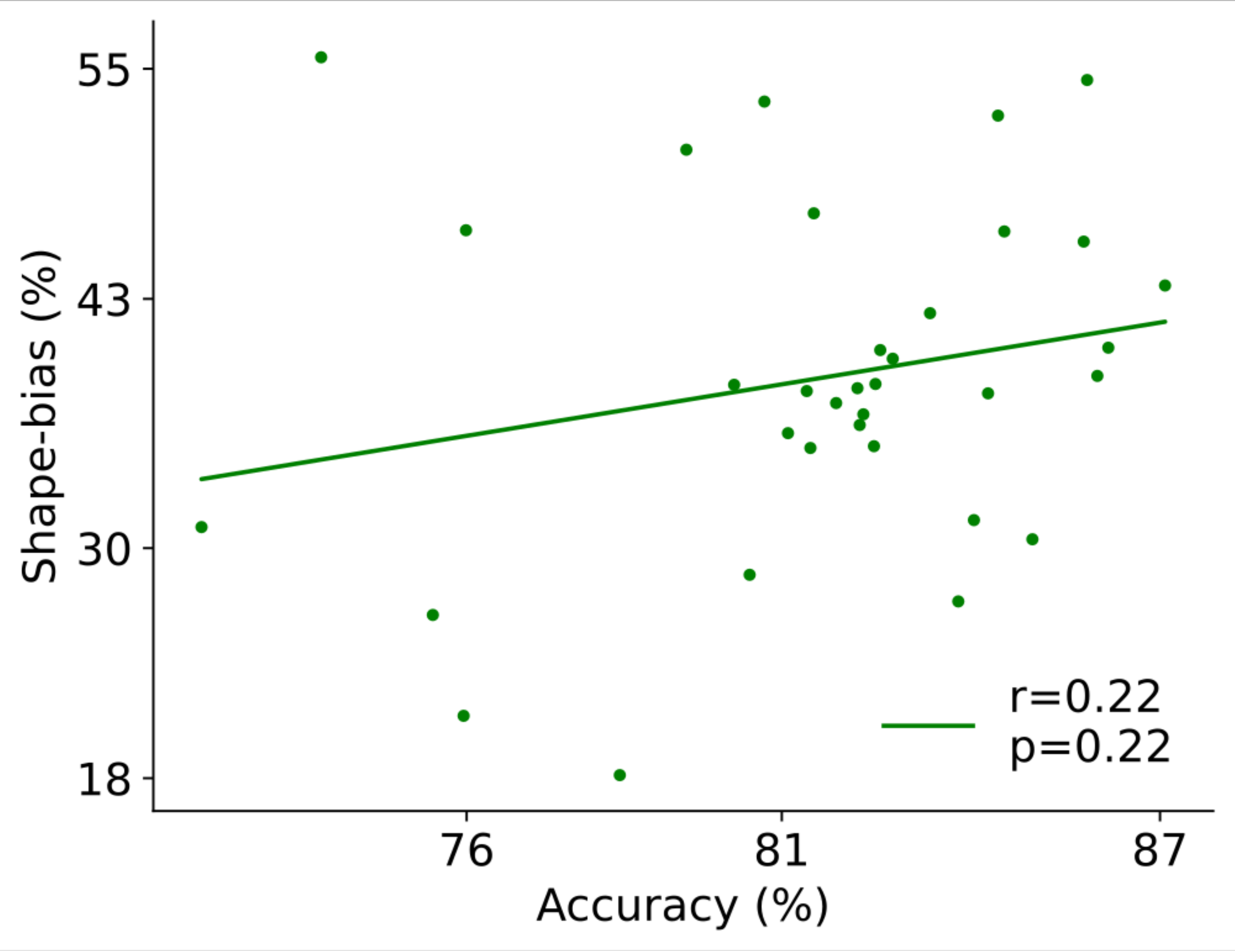}
    \end{subfigure}
    \caption{\textbf{Effect of model family in the correlations between metrics.} Left, distribution of correlation coefficients of different model families for the following metric pairs: accuracy vs shape-bias, accuracy vs shape-dim-ratio, shape-bias vs shape-dim-ratio. Center, example of a model family, ResNext, showing a very strong correlation between shape behavioral bias and ImageNet accuracy. Right, same but for a model, family, ViT, showing no correlation between the same metrics.}
    \label{correlation_distribution}
\end{figure}

How robust are these results for different model families?
We attempt to answer this question by looking at how the metrics under study relate with each other when grouping the models according to their architectural family. We consider all the families that have at least nine different models (16 families in total) and calculate the correlations between the shape behavioral bias, the shape dimensionality bias, and the object classification accuracy (Figure \ref{correlation_distribution}, left). While, the correlation between these metrics is present on most model families (see the correlation between shape-bias and accuracy for the ResNext models on Figure \ref{correlation_distribution}, center; r=0.78, p<10E-9), there is a large degree of variability in how the metrics relate to each other in each model family. For some model families, there is no significant correlation between these metric pairs (see the correlation between shape-bias and accuracy for the ViT models on Figure \ref{correlation_distribution}, center; r=0.22, p<0.22). 

\section{Discussion}

In this paper, we extend the results of previous studies on the relationship between shape bias, measured at both the model's output and penultimate layer, and ImageNet accuracy \cite{hermann2020origins, islam2021shape}. By using a very large and varied model pool, we confirm that optimizing models for performance in an object classification task naturally leads to higher shape bias. This is true even when considering a very wide range of top-1 ImageNet accuracy values ($\sim$60\%-90\%). The best performing model in the model pool is also the one with the highest shape behavioral bias, achieving a level that previously had only been reported with a trade-off in ImageNet accuracy \cite{hermann2020origins}. These are important observations since they suggest that it may not be necessary to adopt strategies to specifically deal with the lack of shape bias in ANNs. If the observed trend continues, one should expect models to continue approaching human-level shape bias as they become increasingly more accurate. When looking into individual model families, while we observe that most follow the overall trend, some have weaker correlations between accuracy and shape bias. The implications of this is that the expected increase in shape bias with increasing ImageNet accuracy may not be present in all ANN architectures. In those cases, additional training strategies should be used to overcome this limitation \cite{hermann2020origins}. Finally, we observe that model improvement in ImageNet accuracy is accompanied by a replacement of texture-based representations by shape-based ones, with residual dimensionality remaining mostly unchanged at their penultimate layer. Thus, there appears to be a trade-off between knowledge about shape and texture at the level of the ANNs' final stages and that texture information is discarded in favor of shape information as models perform better in object classification.

The topic of shape-texture bias in humans and ANNs has received plenty of attention recently, including this study. However, much work remains to be done if one wishes to close the gap between ANNs and human-level shape bias. While we observe that some of the best performing  ANNs have relatively high shape bias ($\sim$60\%), it is still considerably lower than what is observed in humans ($\sim$95\%) \cite{geirhos2018imagenet}. As accuracy in the ImageNet dataset approaches its ceiling, further gains in shape bias may be starting to saturate, and, thus, other optimization goals may be useful. A clear understanding of how shape bias affects other important model performance metrics such as OOD generalizationa and adversarial robustness is also lacking. Future work extending the analyses of this study to include these and other important metrics should provide insights to this topic. Finally, there have been several neuroscience studies looking into how shape and texture information is represented along the primate visual ventral stream areas \cite{cavina2010separate, Kim4760, Freeman2013, hegde_comparative_2007}. It remains to be seen whether ANN models with higher shape bias are also better models of biological vision, particularly in how they represent these two visual concepts.


\label{discussion}


\begin{ack}

This work was supported by the Portuguese Science Foundation, under projects PRELUNA PTDC/CCI-INF/4703/2021 and UIDB/50021/2020. The last author acknowledges the support of a Fulbright fellowship while he was a visiting professor at the CSAIL laboratory at MIT.

\end{ack}

\printbibliography

@article{dapello_simulating_2020,
	title = {Simulating a primary visual cortex at the front of {CNNs} improves robustness to image perturbations},
%	issn = {26928205},
%	doi = {10.1101/2020.06.16.154542},
	abstract = {Current state-of-the-art object recognition models are largely based on convolutional neural network (CNN) architectures, which are loosely inspired by the primate visual system. However, these CNNs can be fooled by imperceptibly small, explicitly crafted perturbations, and struggle to recognize objects in corrupted images that are easily recognized by humans. Here, by making comparisons with primate neural data, we first observed that CNN models with a neural hidden layer that better matches primate primary visual cortex (V1) are also more robust to adversarial attacks. Inspired by this observation, we developed VOneNets, a new class of hybrid CNN vision models. Each VOneNet contains a fixed weight neural network front-end that simulates primate V1, called the VOneBlock, followed by a neural network back-end adapted from current CNN vision models. The VOneBlock is based on a classical neuroscientific model of V1: the linear-nonlinear-Poisson model, consisting of a biologically-constrained Gabor filter bank, simple and complex cell nonlinearities, and a V1 neuronal stochasticity generator. After training, VOneNets retain high ImageNet performance, but each is substantially more robust, outperforming the base CNNs and state-of-the-art methods by 18\% and 3\%, respectively, on a conglomerate benchmark of perturbations comprised of white box adversarial attacks and common image corruptions. Finally, we show that all components of the VOneBlock work in synergy to improve robustness. While current CNN architectures are arguably brain-inspired, the results presented here demonstrate that more precisely mimicking just one stage of the primate visual system leads to new gains in ImageNet-level computer vision applications.},
	journal = {NeurIPS},
	author = {Dapello, Joel and Marques, Tiago and Schrimpf, Martin and Geiger, Franziska and Cox, David D. and DiCarlo, James J.},
	year = {2020},
	pages = {1--30},
	file = {PDF:/Users/tmarques/Zotero/storage/NSDYWMZI/Dapello et al._2020(3).pdf:application/pdf},
}

@article{hendrycks_benchmarking_2019,
	title = {Benchmarking neural network robustness to common corruptions and perturbations},
	journal = {7th International Conference on Learning Representations, ICLR 2019},
	author = {Hendrycks, Dan and Dietterich, Thomas},
	year = {2019},
	pages = {1--16},
}

@article{hermann2020origins,
  title={The origins and prevalence of texture bias in convolutional neural networks},
  author={Hermann, Katherine and Chen, Ting and Kornblith, Simon},
  journal={Advances in Neural Information Processing Systems},
  volume={33},
  pages={19000--19015},
  year={2020}
}

@article{islam2021shape,
  title={Shape or texture: Understanding discriminative features in {CNN}s},
  author={Islam, Md Amirul and Kowal, Matthew and Esser, Patrick and Jia, Sen and Ommer, Bjorn and Derpanis, Konstantinos G and Bruce, Neil},
  journal={arXiv preprint arXiv:2101.11604},
  year={2021}
}

@article{geirhos2018imagenet,
  title={{ImageNet}-trained {CNN}s are biased towards texture; increasing shape bias improves accuracy and robustness},
  author={Geirhos, Robert and Rubisch, Patricia and Michaelis, Claudio and Bethge, Matthias and Wichmann, Felix A and Brendel, Wieland},
  journal={arXiv preprint arXiv:1811.12231},
  year={2018}
}

@article{geirhos2021partial,
  title={Partial success in closing the gap between human and machine vision},
  author={Geirhos, Robert and Narayanappa, Kantharaju and Mitzkus, Benjamin and Thieringer, Tizian and Bethge, Matthias and Wichmann, Felix A and Brendel, Wieland},
  journal={Advances in Neural Information Processing Systems},
  volume={34},
  pages={23885--23899},
  year={2021}
}

@inproceedings{szegedy2014intriguing,
  title={Intriguing properties of neural networks},
  author={Szegedy, Christian and Zaremba, Wojciech and Sutskever, Ilya and Bruna, Joan and Erhan, Dumitru and Goodfellow, Ian and Fergus, Rob},
  booktitle={2nd International Conference on Learning Representations, ICLR 2014},
  year={2014}
}

@incollection{kurakin2018adversarial,
  title={Adversarial examples in the physical world},
  author={Kurakin, Alexey and Goodfellow, Ian J and Bengio, Samy},
  booktitle={Artificial intelligence safety and security},
  pages={99--112},
  year={2018},
  publisher={Chapman and Hall/CRC}
}

@article{chen2020shape,
  title={The shape and simplicity biases of adversarially robust ImageNet-trained CNNs},
  author={Chen, Peijie and Agarwal, Chirag and Nguyen, Anh},
  journal={arXiv preprint arXiv:2006.09373},
  year={2020}
}

@inproceedings{zhang2019interpreting,
  title={Interpreting adversarially trained convolutional neural networks},
  author={Zhang, Tianyuan and Zhu, Zhanxing},
  booktitle={International Conference on Machine Learning},
  pages={7502--7511},
  year={2019},
  organization={PMLR}
}

@article{geirhos2018generalisation,
  title={Generalisation in humans and deep neural networks},
  author={Geirhos, Robert and Temme, Carlos RM and Rauber, Jonas and Sch{\"u}tt, Heiko H and Bethge, Matthias and Wichmann, Felix A},
  journal={Advances in neural information processing systems},
  volume={31},
  year={2018}
}

@article{mummadi2021does,
  title={Does enhanced shape bias improve neural network robustness to common corruptions?},
  author={Mummadi, Chaithanya Kumar and Subramaniam, Ranjitha and Hutmacher, Robin and Vitay, Julien and Fischer, Volker and Metzen, Jan Hendrik},
  journal={arXiv preprint arXiv:2104.09789},
  year={2021}
}

@inproceedings{xie2020adversarial,
  title={Adversarial examples improve image recognition},
  author={Xie, Cihang and Tan, Mingxing and Gong, Boqing and Wang, Jiang and Yuille, Alan L and Le, Quoc V},
  booktitle={Proceedings of the IEEE/CVF Conference on Computer Vision and Pattern Recognition},
  pages={819--828},
  year={2020}
}

@inproceedings{gatys2016image,
  title={Image style transfer using convolutional neural networks},
  author={Gatys, Leon A and Ecker, Alexander S and Bethge, Matthias},
  booktitle={Proceedings of the IEEE conference on computer vision and pattern recognition},
  pages={2414--2423},
  year={2016}
}

@article{zhuang2021unsupervised,
  title={Unsupervised neural network models of the ventral visual stream},
  author={Zhuang, Chengxu and Yan, Siming and Nayebi, Aran and Schrimpf, Martin and Frank, Michael C and DiCarlo, James J and Yamins, Daniel LK},
  journal={Proceedings of the National Academy of Sciences},
  volume={118},
  number={3},
  pages={e2014196118},
  year={2021},
}

@article{cavina2010separate,
  title={Separate processing of texture and form in the ventral stream: evidence from FMRI and visual agnosia},
  author={Cavina-Pratesi, C and Kentridge, RW and Heywood, CA and Milner, AD},
  journal={Cerebral Cortex},
  volume={20},
  number={2},
  pages={433--446},
  year={2010},
  publisher={Oxford University Press}
}

@inproceedings{xie2020self,
  title={Self-training with noisy student improves {ImageNet} classification},
  author={Xie, Qizhe and Luong, Minh-Thang and Hovy, Eduard and Le, Quoc V},
  booktitle={Proceedings of the IEEE/CVF conference on computer vision and pattern recognition},
  pages={10687--10698},
  year={2020}
}

@inproceedings{he2016deep,
  title={Deep residual learning for image recognition},
  author={He, Kaiming and Zhang, Xiangyu and Ren, Shaoqing and Sun, Jian},
  booktitle={Proceedings of the IEEE conference on computer vision and pattern recognition},
  pages={770--778},
  year={2016}
}

@article{ali2021xcit,
  title={{XCiT}: Cross-covariance image transformers},
  author={Ali, Alaaeldin and Touvron, Hugo and Caron, Mathilde and Bojanowski, Piotr and Douze, Matthijs and Joulin, Armand and Laptev, Ivan and Neverova, Natalia and Synnaeve, Gabriel and Verbeek, Jakob and others},
  journal={Advances in neural information processing systems},
  volume={34},
  pages={20014--20027},
  year={2021}
}

@inproceedings{liu2022convnet,
  title={A {ConvNet} for the 2020s},
  author={Liu, Zhuang and Mao, Hanzi and Wu, Chao-Yuan and Feichtenhofer, Christoph and Darrell, Trevor and Xie, Saining},
  booktitle={Proceedings of the IEEE/CVF Conference on Computer Vision and Pattern Recognition},
  pages={11976--11986},
  year={2022}
}

@article{wang2020deep,
  title={Deep high-resolution representation learning for visual recognition},
  author={Wang, Jingdong and Sun, Ke and Cheng, Tianheng and Jiang, Borui and Deng, Chaorui and Zhao, Yang and Liu, Dong and Mu, Yadong and Tan, Mingkui and Wang, Xinggang and others},
  journal={IEEE transactions on pattern analysis and machine intelligence},
  volume={43},
  number={10},
  pages={3349--3364},
  year={2020},
  publisher={IEEE}
}

@article{touvron2021resmlp,
  title={{ResMLP}: Feedforward networks for image classification with data-efficient training},
  author={Touvron, Hugo and Bojanowski, Piotr and Caron, Mathilde and Cord, Matthieu and El-Nouby, Alaaeldin and Grave, Edouard and Izacard, Gautier and Joulin, Armand and Synnaeve, Gabriel and Verbeek, Jakob and others},
  journal={arXiv preprint arXiv:2105.03404},
  year={2021}
}

@inproceedings{tan2021efficientnetv2,
  title={{EfficientNetV2}: Smaller models and faster training},
  author={Tan, Mingxing and Le, Quoc},
  booktitle={International Conference on Machine Learning},
  pages={10096--10106},
  year={2021},
  organization={PMLR}
}

@inproceedings{tan2019efficientnet,
  title={{EfficientNet}: Rethinking model scaling for convolutional neural networks},
  author={Tan, Mingxing and Le, Quoc},
  booktitle={International conference on machine learning},
  pages={6105--6114},
  year={2019},
  organization={PMLR}
}

@article{yuan2021volo,
  title={{VOLO}: Vision outlooker for visual recognition},
  author={Yuan, Li and Hou, Qibin and Jiang, Zihang and Feng, Jiashi and Yan, Shuicheng},
  journal={arXiv preprint arXiv:2106.13112},
  year={2021}
}

@inproceedings{liu2022swin,
  title={Swin transformer {V2}: Scaling up capacity and resolution},
  author={Liu, Ze and Hu, Han and Lin, Yutong and Yao, Zhuliang and Xie, Zhenda and Wei, Yixuan and Ning, Jia and Cao, Yue and Zhang, Zheng and Dong, Li and others},
  booktitle={Proceedings of the IEEE/CVF Conference on Computer Vision and Pattern Recognition},
  pages={12009--12019},
  year={2022}
}

@inproceedings{liu2021swin,
  title={Swin transformer: Hierarchical vision transformer using shifted windows},
  author={Liu, Ze and Lin, Yutong and Cao, Yue and Hu, Han and Wei, Yixuan and Zhang, Zheng and Lin, Stephen and Guo, Baining},
  booktitle={Proceedings of the IEEE/CVF International Conference on Computer Vision},
  pages={10012--10022},
  year={2021}
}

@article{xu2022regnet,
  title={RegNet: self-regulated network for image classification},
  author={Xu, Jing and Pan, Yu and Pan, Xinglin and Hoi, Steven and Yi, Zhang and Xu, Zenglin},
  journal={IEEE Transactions on Neural Networks and Learning Systems},
  year={2022},
  publisher={IEEE}
}

@article{mehta2021mobilevit,
  title={{MobileViT}: light-weight, general-purpose, and mobile-friendly vision transformer},
  author={Mehta, Sachin and Rastegari, Mohammad},
  journal={arXiv preprint arXiv:2110.02178},
  year={2021}
}

@article{howard2017mobilenets,
  title={{MobileNets}: Efficient convolutional neural networks for mobile vision applications},
  author={Howard, Andrew G and Zhu, Menglong and Chen, Bo and Kalenichenko, Dmitry and Wang, Weijun and Weyand, Tobias and Andreetto, Marco and Adam, Hartwig},
  journal={arXiv preprint arXiv:1704.04861},
  year={2017}
}

@inproceedings{sandler2018mobilenetv2,
  title={{MobileNetV2}: Inverted residuals and linear bottlenecks},
  author={Sandler, Mark and Howard, Andrew and Zhu, Menglong and Zhmoginov, Andrey and Chen, Liang-Chieh},
  booktitle={Proceedings of the IEEE conference on computer vision and pattern recognition},
  pages={4510--4520},
  year={2018}
}

@inproceedings{howard2019searching,
  title={Searching for {MobileNetV3}},
  author={Howard, Andrew and Sandler, Mark and Chu, Grace and Chen, Liang-Chieh and Chen, Bo and Tan, Mingxing and Wang, Weijun and Zhu, Yukun and Pang, Ruoming and Vasudevan, Vijay and others},
  booktitle={Proceedings of the IEEE/CVF international conference on computer vision},
  pages={1314--1324},
  year={2019}
}

@inproceedings{touvron2021training,
  title={Training data-efficient image transformers \& distillation through attention},
  author={Touvron, Hugo and Cord, Matthieu and Douze, Matthijs and Massa, Francisco and Sablayrolles, Alexandre and J{\'e}gou, Herv{\'e}},
  booktitle={International Conference on Machine Learning},
  pages={10347--10357},
  year={2021},
  organization={PMLR}
}

@inproceedings{chen2021crossvit,
  title={{CrossViT}: Cross-attention multi-scale vision transformer for image classification},
  author={Chen, Chun-Fu Richard and Fan, Quanfu and Panda, Rameswar},
  booktitle={Proceedings of the IEEE/CVF international conference on computer vision},
  pages={357--366},
  year={2021}
}

@inproceedings{touvron2021going,
  title={Going deeper with image transformers},
  author={Touvron, Hugo and Cord, Matthieu and Sablayrolles, Alexandre and Synnaeve, Gabriel and J{\'e}gou, Herv{\'e}},
  booktitle={Proceedings of the IEEE/CVF International Conference on Computer Vision},
  pages={32--42},
  year={2021}
}

@inproceedings{esser2020disentangling,
  title={A disentangling invertible interpretation network for explaining latent representations},
  author={Esser, Patrick and Rombach, Robin and Ommer, Bjorn},
  booktitle={Proceedings of the IEEE/CVF Conference on Computer Vision and Pattern Recognition},
  pages={9223--9232},
  year={2020}
}

@article{everingham2010pascal,
  title={The {PASCAL} visual object classes ({VOC}) challenge},
  author={Everingham, Mark and Van Gool, Luc and Williams, Christopher KI and Winn, John and Zisserman, Andrew},
  journal={International journal of computer vision},
  volume={88},
  number={2},
  pages={303--338},
  year={2010},
  publisher={Springer}
}

@inproceedings{cimpoi2014describing,
  title={Describing textures in the wild},
  author={Cimpoi, Mircea and Maji, Subhransu and Kokkinos, Iasonas and Mohamed, Sammy and Vedaldi, Andrea},
  booktitle={Proceedings of the IEEE conference on computer vision and pattern recognition},
  pages={3606--3613},
  year={2014}
}

@article{dosovitskiy2020image,
  title={An image is worth 16x16 words: Transformers for image recognition at scale},
  author={Dosovitskiy, Alexey and Beyer, Lucas and Kolesnikov, Alexander and Weissenborn, Dirk and Zhai, Xiaohua and Unterthiner, Thomas and Dehghani, Mostafa and Minderer, Matthias and Heigold, Georg and Gelly, Sylvain and others},
  journal={arXiv preprint arXiv:2010.11929},
  year={2020}
}

@article {Kim4760,
	author = {Kim, Taekjun and Bair, Wyeth and Pasupathy, Anitha},
	title = {Neural Coding for Shape and Texture in Macaque Area V4},
	volume = {39},
	number = {24},
	pages = {4760--4774},
	year = {2019},
	publisher = {Society for Neuroscience},
	journal = {Journal of Neuroscience}
}

@article{Freeman2013,
	title = {A functional and perceptual signature of the second visual area in primates},
	volume = {16},
	number = {7},
	journal = {Nature Neuroscience},
	author = {Freeman, Jeremy and Ziemba, Corey M. and Heeger, David J. and Simoncelli, E. P. and Movshon, J. A.},
	year = {2013},
	pmid = {23685719},
	pages = {974--981},
}

@article{hegde_comparative_2007,
	title = {A comparative study of shape representation in macaque visual areas {V2} and {V4}},
	volume = {17},
	number = {5},
	journal = {Cerebral Cortex},
	author = {Hegdé, Jay and Van Essen, David C.},
	year = {2007},
	pages = {1100--1116},
}





\end{document}